\RequirePackage{amsmath}
\documentclass[runningheads]{llncs}
\usepackage{graphicx}
\usepackage[caption=false]{subfig} 

\usepackage{graphicx}
\usepackage{amsmath} 
\usepackage{mathtools}
\usepackage{tikz}
\usetikzlibrary{shapes.geometric, arrows}
\usetikzlibrary{fit,positioning,arrows}
\usepackage{multirow}
\usepackage{lscape}
\usepackage[normalem]{ulem}
\useunder{\uline}{\ul}{}
\usepackage{cite}

\usepackage{booktabs}
\usepackage{makecell}
\usepackage{tabularx}
\usepackage{rotating}
\usepackage[export]{adjustbox}

\newlength{\bibitemsep}
\newlength{\bibparskip}
\let\oldthebibliography\thebibliography
\renewcommand\thebibliography[1]{%
  \oldthebibliography{#1}%
  \setlength{\parskip}{\bibitemsep}%
  \setlength{\itemsep}{\bibparskip}%
}

\begin{document}
\title{Supervised and Unsupervised Detections for Multiple Object Tracking in Traffic Scenes: A Comparative Study}
%
\titlerunning{Supervised and Unsupervised Detections for MOT}
%
\author{Hui-Lee Ooi\inst{1}
\and
Guillaume-Alexandre Bilodeau\inst{1} \and
Nicolas Saunier\inst{2}}
\authorrunning{Ooi et al.}
%
\institute{LITIV, Department of Computer and Software Engineering, Polytechnique Montr\'{e}al, Canada \and
Department of Civil, Geological and Mining Engineering, Polytechnique Montr\'{e}al, Canada
}
\maketitle              
\begin{abstract}
In this paper, we propose a multiple object tracker, called MF-Tracker, that integrates multiple classical features (spatial distances and colours) and modern features (detection labels and re-identification features) in its tracking framework. Since our tracker can work with detections coming either from unsupervised and supervised object detectors, we also investigated the impact of supervised and unsupervised detection inputs in our method and for tracking road users in general. We also compared our results with existing methods that were applied on the UA-Detrac and the UrbanTracker datasets. Results show that our proposed method is performing very well in both datasets with different inputs (MOTA ranging from $0.3491$ to $0.5805$ for unsupervised inputs on the UrbanTracker dataset and an average MOTA of $0.7638$ for supervised inputs on the UA Detrac dataset) under different circumstances. A well-trained supervised object detector can give better results in challenging scenarios. However, in simpler scenarios, if good training data is not available, unsupervised method can perform well and can be a good alternative.

\end{abstract}
\section{Introduction}
\begin{figure*}[t!]
    \centering
    \subfloat[]{
      \includegraphics[width=0.18\paperwidth]{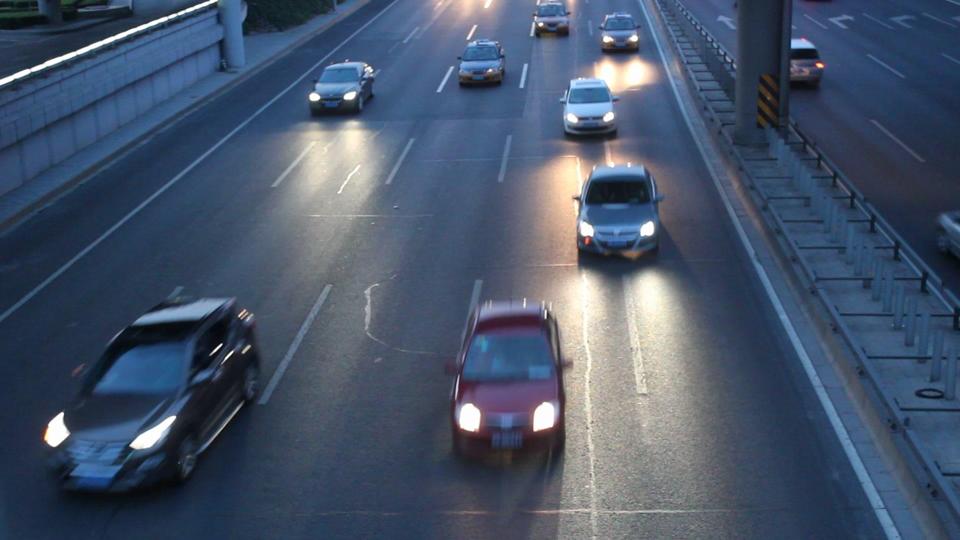}}
~
     \subfloat[]{
      \includegraphics[width=0.18\paperwidth]{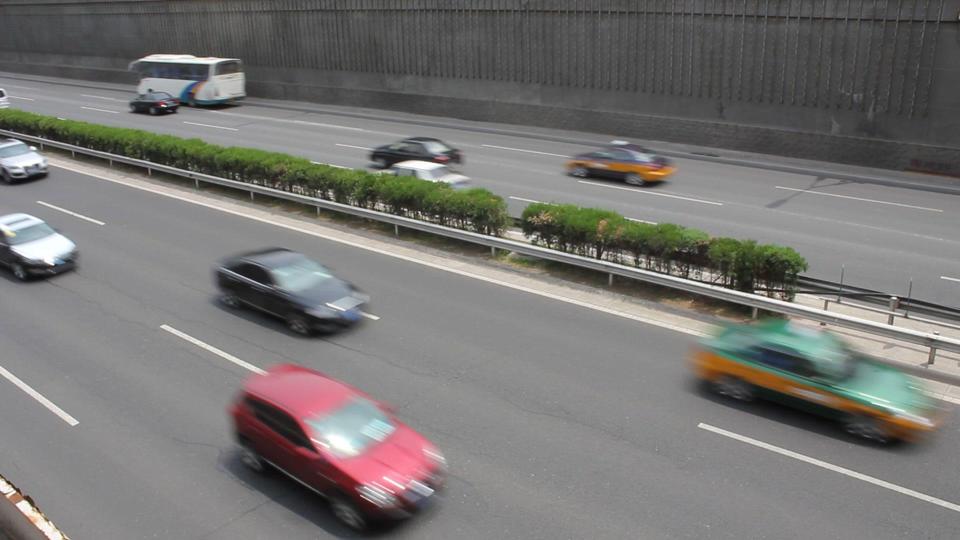}}
~   
    \subfloat[]{
      \includegraphics[width=0.18\paperwidth]{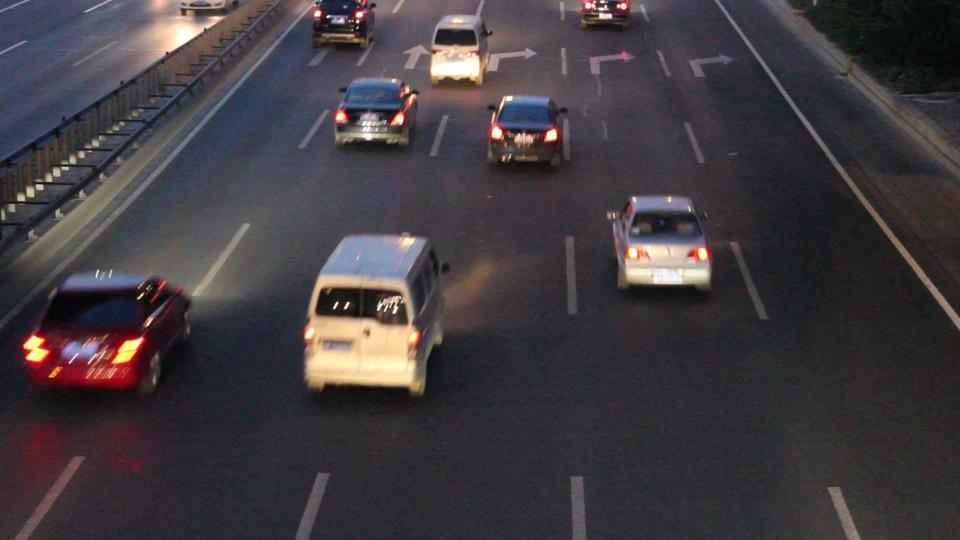}}
    \caption{Examples of selected frames from videos in the UA-Detrac dataset~\cite{wen2015ua} used for evaluation in the experiments.}
\label{fig_eg}
\end{figure*}

Multiple object tracking (MOT) in the context of traffic scenes essentially means following the target objects (road users) in the scene to obtain an accurate representation of their trajectories across frames, usually as feedback information to eventually improve traffic management systems or to better plan the layout of the roads. To follow an object, we must see it first; to track a road user in a scene, the importance of getting correct detection inputs for the tracking paradigm must not be overlooked. Compared to single object tracking, MOT has to keep track of the presence of more than one target object while dealing with the possible occlusions and mismatches of objects as a result of interactions of the moving objects with the background and other objects, making it a challenging problem that is still actively researched. In the case of traffic scenes, the MOT method must also deal with various lighting and weather conditions (See Figure \ref{fig_eg}). There are also multiple classes of objects. 

Generally, there are two types of object detection methods to be used for tracking: supervised and unsupervised. The former is the more modern approach using labeled data to train models that can detect the target objects in a particular domain~\cite{lin2017focal,wang2017evolving}. This approach usually delineates an object with a bounding box, and also attributes a class label to each detected object. The latter typically corresponds to the classical approach of foreground extraction and outputs objects that are not part of the background in the frame~\cite{barnich2010vibe,st2016universal}. This method does not need supervised training as it segments the scene in two classes based on a model of the background. It is designed for cameras that are not moving and provides an object segmentation mask, but no labels.

In this paper, we address the MOT problem for traffic scenes by proposing a new tracker that integrates classical features (spatial distances and colours) and modern features (detection labels and re-identification features), as well as object prediction in its tracking framework. Our tracker can be applied to either supervised and unsupervised object detections. Therefore, while designing our method we raised the question: \emph{which type of detection should be used?} To answer this question, we investigated formally the impact of the choice of the type of detections in the design of the tracker.

The contributions of this paper are: 1) a new MOT tracker that combines various features and that can capitalize on both unsupervised and supervised object detections and 2) a formal analysis of the performance of unsupervised and supervised object detectors in road tracking scenarios and their impact on MOT.

\section{Related Works}
The study of MOT on traffic scenes has undergone many changes and evolution over the years. Conventionally, before the advent of deep learning in computer vision, the extraction of target objects in the application of MOT were generic and unsupervised, as in \cite{breitenstein2010online, jodoin2016tracking, yang2017multiple}. In Yang et al. \cite{yang2017multiple}, background subtraction detections were combined with kernelized correlation filters (KCF) for solving the MOT problem in urban traffic scenes. KCF is used as an appearance model as well as for predicting the object position in the next frame. Similarly, Jodoin et al. \cite{jodoin2016tracking} used background subtraction to extract potential unknown road users for their proposed finite state machine to handle the different target objects. Keypoints are used as an appearance model. Other works, like the one of Saunier et al. \cite{saunier2006feature}, instead used optical flow information to detect and track road users. 

Recently, most works on MOT use detections from supervised learning methods that output bounding boxes around learned object classes. The use of a deep learning-based detector as the only source of input for multiple object tracking involving several different types of road users was presented in~\cite{ooi2018multiple}, but with disappointing results. Ooi et al. \cite{ooi2019tracking} then further improved the method on the same dataset (UrbanTracker~\cite{jodoin2016tracking}) by applying classical unsupervised object detection outputs coupled with modern supervised learning-based detector outputs, achieving some progress with the use of detector labels as part of the feature description as well. 

Meanwhile, the reported results on the UA-Detrac dataset \cite{wen2015ua} on its official website are based on supervised object detectors. UA-Detrac does not consider bikes, motorcycles and pedestrians. At the time of conducting our experiments, the reported top trackers on the dataset are Evolving Boxes (EB)+Kalman+IOUT (extension of \cite{bochinski2017high}), EB+IOUT \cite{bochinski2017high} and RCNN+IOUT \cite{bochinski2017high}. These three methods are rather similar, essentially working by the overlap of the intersection over union (IOU) of the bounding boxes that represent the objects in each frame, with the assumptions that the high frame rate of the videos does not leave ``gaps'' between the detections~\cite{bochinski2017high}. The Kalman filter used in the EB+Kalman+IOUT approach is meant to allow skipping frames via predictions to improve processing speed. 

Hence, in this study, we are interested in evaluating and understanding the effects of unsupervised and supervised detections for MOT in varying traffic scenarios under different environmental conditions as provided by these two datasets, UA-Detrac and UrbanTracker. We therefore devised a novel tracker that can work with both kinds of inputs.

\section{Method}
We proposed a novel tracker (MF-Tracker) that combines classical features as well as deep learning features for the matching of objects across frames. We are also interested in investigating the effects of supervised and unsupervised detections on MOT performance. Our tracker was thus designed to work with both types of detections. 

\begin{figure*}[t!]
\centering
\tikzstyle{inputbox} = [rectangle, rounded corners, minimum width=2.5cm, minimum height=1cm,text centered, text width=1.8cm, draw=black, fill=none]
\tikzstyle{computebox} = [rectangle, minimum width=2cm, minimum height=3cm,text centered, text width=1.8cm, draw=black, fill=none]
\tikzstyle{nobox} = [rectangle, text centered, text width=1.8cm, fill=none]

\tikzstyle{arrow} = [thick,->,>=stealth]
\tikzstyle{bidir}= [thick,<->,>=stealth]

\begin{tikzpicture}[node distance=2cm]
\node (supervised) [inputbox]{Supervised detections};
\node (unsupervised) [inputbox, below of=supervised, yshift=-0.1cm] {Unsupervised detections};
\node (or) [nobox, above of=unsupervised, yshift=-1cm] {OR};
\node (extraction) [computebox, right of=supervised, xshift=1cm, yshift=-1cm] {Feature Extraction};
\node (association) [computebox, right of=extraction, xshift=1.7cm] {Data Association};
\node (Track) [inputbox, right of=association, xshift=1cm] {Tracks};
\node (feat1) [nobox, left of=extraction, xshift=3.8cm, yshift=1cm] {spatial};
\node (feat2) [nobox, left of=extraction, xshift=3.8cm, yshift=0.3cm] {color};
\node (feat2) [nobox, left of=extraction, xshift=3.8cm, yshift=-0.4cm] {label};
\node (feat4) [nobox, left of=extraction, xshift=3.8cm, yshift=-1cm] {REID};

\draw [arrow, dotted] (supervised) -- (extraction);
\draw [arrow,dotted] (unsupervised) -- (extraction);
\draw [arrow] (association) -- (Track);
\draw [arrow](extraction.east) -- (association.west);
\draw [arrow]([yshift=0.8 cm]extraction.east) -- ([yshift=0.8 cm]association.west);
\draw [arrow]([yshift=-0.8 cm]extraction.east) -- ([yshift=-0.8 cm]association.west);
\draw [arrow]([yshift=-1.3 cm]extraction.east) -- ([yshift=-1.3 cm]association.west);

\end{tikzpicture}
\caption{Overview of our proposed tracker (MF-Tracker). Detections from supervised or unsupervised approaches are fed into the Feature Extraction module for further processing in Data Association to produce the final trajectory outputs.} 
\label{fig_flowchart}
\end{figure*}
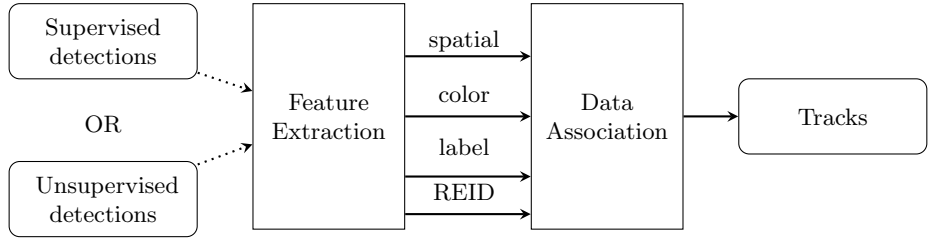

Our multiple object tracker consists of several components: (i) Object detection, (ii) Feature generation from objects and (iii) Data association to produce the final tracking outputs that describe the trajectory of each target object across frames, as shown in Figure~\ref{fig_flowchart}.

\subsection{Inputs for the Tracker}
Since we intend to compare the performances of different input objects for tracking, we used a state-of-the-art background subtraction method (IMOT \cite{beaupre2018improving} with PAWCS (Pixel-based Adaptive Word Consensus Segmenter) \cite{st2016universal}) as unsupervised input source and the deep learning-based detector (RetinaNet \cite{lin2017focal}) as supervised input source. Both approaches give bounding boxes of target objects for each frame. 

The next step for MF-Tracker is to extract the information contained within the bounding boxes for the subsequent tracking module.

\subsection{Classical Features and Modern Features}
The proposed method integrates both classical features and modern features to generate overall similarity scores to compare the objects across frames. 

The similarity costs from classical features are:
\begin{itemize}
    \item the spatial cost $C_d$: based on the spatial distances of the four coordinates of the bounding boxes, it is defined as:
    
\begin{equation}
C_{d} = 1 - max (0, \frac{T_{d} - \overline{SD}}{T_{d}}) 
\label{eqn:cost_spatial}
\end{equation}

\begin{multline}
\overline{SD} =\frac{1}{4}(|x_{D,min} - x_{T,min}| + |y_{D,min} - y_{T,min}| + \\ |x_{D,max} - x_{T,max}| + |y_{D,max} - y_{T,max}|),
\label{eqn:distance_avg}
\end{multline}
where $\overline{SD}$ is the mean bounding box spatial distance and  $x_{min}$, $y_{min}$, $x_{max}$ and $y_{max}$ denote the minimum and maximum coordinates of an object bounding box. $T$ represents an object that is currently tracked while $D$ represents a detected object in a frame. $\overline{SD}$ denotes the mean spatial distance of the $x$ coordinates and $y$ coordinates of all the four corners of the bounding boxes of the compared objects. A fixed parameter $T_{d}$ is used to normalize $C_{d}$ and to bound the maximal distance between bounding boxes.
    
    \item the color cost $C_c$: it is the Bhattacharyya distances of the color histograms of the bounding boxes. It is defined as: 
\begin{equation}
  C_{c} =\sqrt[]{1-\frac{1}{\sqrt[]{\overline{H^D_{i}}\overline{H^T_{j}}N^2}}\sum_N{\sqrt[]{H^D_i H^T_j}}}, 
  \label{eqn:cost_color}
\end{equation}
where $H^D_{i}$ denotes the color histogram of a detection $i$, $H^T_{j}$ denotes the color histogram of a currently tracked object $j$ and $N$ is the total number of histogram bins (256 is used in this work). $\overline{H^D_{m}}$ and $\overline{H^T_{m}}$ are the histogram bin means of the detected object and currently tracked object, given by Equation \ref{eqn:mean_bhat_d}
and Equation \ref{eqn:mean_bhat_t} respectively. 
\begin{equation}
\overline{H^D_{m}} =\frac{1}{N}\sum_{}^{} H^D_{m}
\label{eqn:mean_bhat_d}
\end{equation}

\begin{equation}
\overline{H^T_{m}} =\frac{1}{N}\sum_{}^{} H^T_{m}
\label{eqn:mean_bhat_t}
\end{equation}
\end{itemize}

Meanwhile, the similarity costs from modern features are:
\begin{itemize}
    \item the label cost $C_l$: the label information from the detector inputs are used as a similarity cost. It is defined as:
    \begin{equation}
  C_{l}=
  \begin{cases}
    1 - \frac{W_{i} + W_{j}}{2} & \text{if $L_{i}=L_{j}$ } \\
    1 & \text{if $L_{i} \neq L_{j}$}, 
  \end{cases}
\label{eqn:cost_label}
\end{equation}
where $L_{i}$ denotes the class label of object $i$ and $W_{i}$ its confidence value (between 0 and 1). Using the confidence value from the object class label, and not just the class label for the cost is a beneficial strategy because confidence values tend to be similar in consecutive frames for a given object. 
    \item the re-identification (REID) cost $C_r$: the deep-learned REID features of OSNet~\cite{zhou2019omni} are also used for object description, where the REID cost is computed with the Euclidean distance as 
\begin{equation}
C_{r}= 1 - \sqrt[] {\sum_n{(r^i_n-r^j_n)^2}}
\label{eqn:cost_reid}
\end{equation}

where $r^i_n$ and $r^j_n$ denote respectively the $n_{th}$ REID feature value of object $i$ and $j$, and $n$ is the number of REID features. We used OSNet pre-trained on Multi-Scene Multi-Time person ReID dataset (MSMT17)~\cite{wei2018person}. The features were not specifically tuned for our application.  
\end{itemize}

All these features are applied and combined to give a final similarity score given by 

\begin{equation}
  C_{final}= \alpha C_{d} + \beta C_{c} + \gamma C_{l} + \lambda C_{r},
\label{eqn:cost_final}
\end{equation}
that ranges from 0 to 1, and where $\alpha, \beta, \gamma, \lambda$ denotes the weights for the corresponding cost.

This procedure is performed in the extracted bounding boxes of detections from both supervised and unsupervised sources. 
In the experiment, for the case of unsupervised detections, due to lack of label information from the unsupervised method itself, detections from the supervised detector are matched with the ones from the unsupervised approach, thus assigning the label accordingly to the bounding boxes given by the unsupervised detector. Alternatively, an object classifier could be applied. Input detection boxes from the unsupervised approach are given null labels if there is no overlapping boxes from the supervised approach.  

\subsection{Data Association}
Based on the similarity score computed from the features, the Hungarian algorithm is used for matching the detected objects (detection list) from the supervised or unsupervised approaches in each frame to the tracked objects (tracked list) accumulated from the previous frames. 

Corresponding objects from the two lists (detection list and tracked list) are marked as matched detection and the information for the objects is updated accordingly. Objects from the detection list that are not successfully matched with any object in the tracked list are initialized as new objects and taken in as part of the tracked list for the subsequent frame. Unmatched objects from the tracked list could either be objects that are occluded or objects that already left the scenes, or invalid objects that are incorrectly detected. A Kalman filter is used to make prediction in the subsequent frames, accounting for occlusion cases, so that occluded objects in the tracked list proceed with possible trajectories when they were momentarily not detected at certain frames. 
 
For each object trajectory, there is also an analysis on the position histories so as to remove invalid objects that are not relevant or to terminate the trajectory when the objects were confirmed to have left the scene.

\section{Experiments}

\begin{figure}[t!]
\centering
  \begin{tabular}{ *{5}{c} }
    ~\rotatebox[origin=c]{90}{\smash{Supervised}}~~
        \includegraphics[valign=c,width=0.2\linewidth]{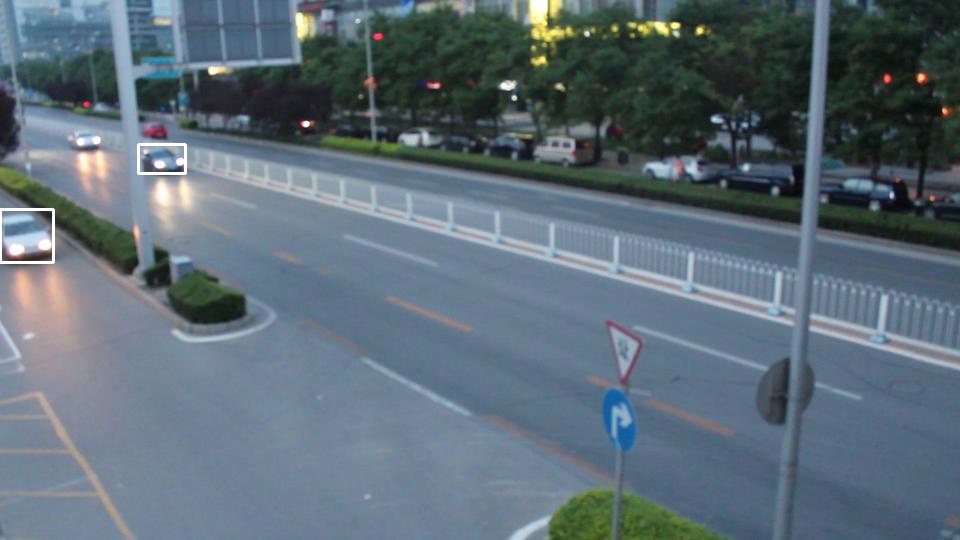}
      & \includegraphics[valign=c,width=0.2\linewidth]{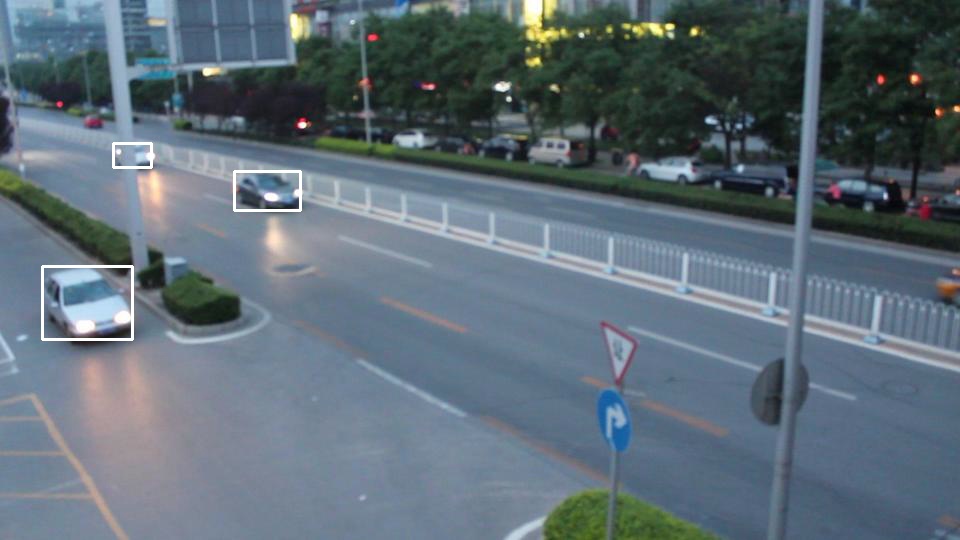}
      & \includegraphics[valign=c,width=0.2\linewidth]{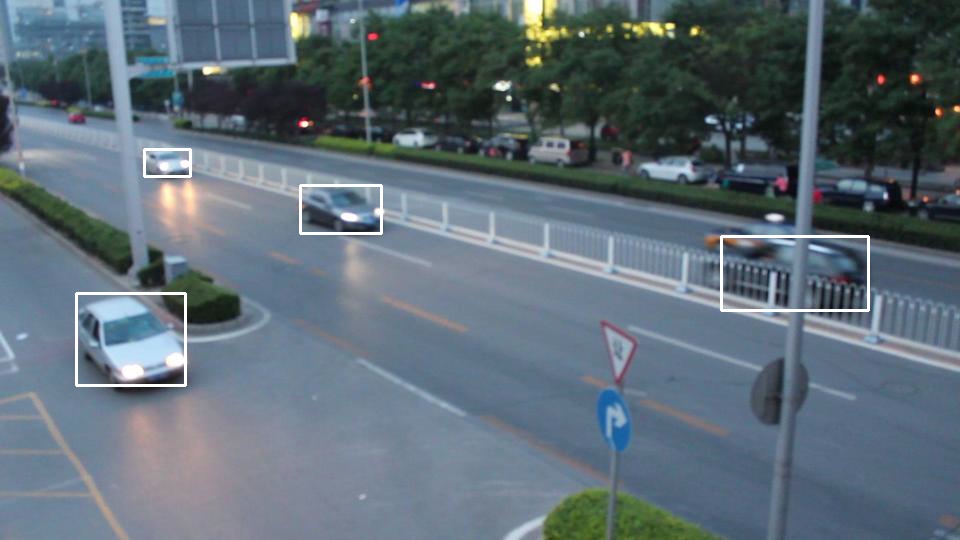}
      & \includegraphics[valign=c,width=0.2\linewidth]{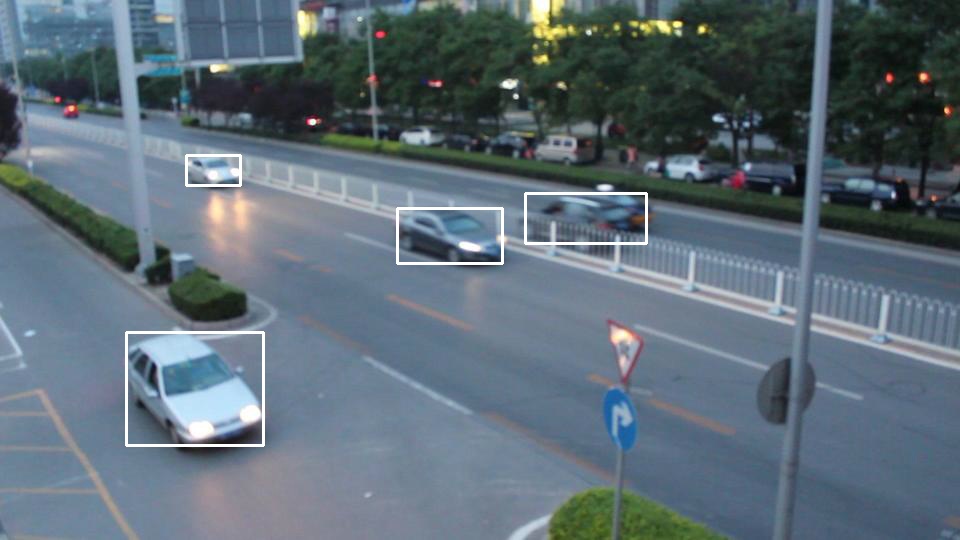} \\ \\
    ~\rotatebox[origin=c]{90}{\smash{Unsupervised}}~~
        \includegraphics[valign=c,width=0.2\linewidth]{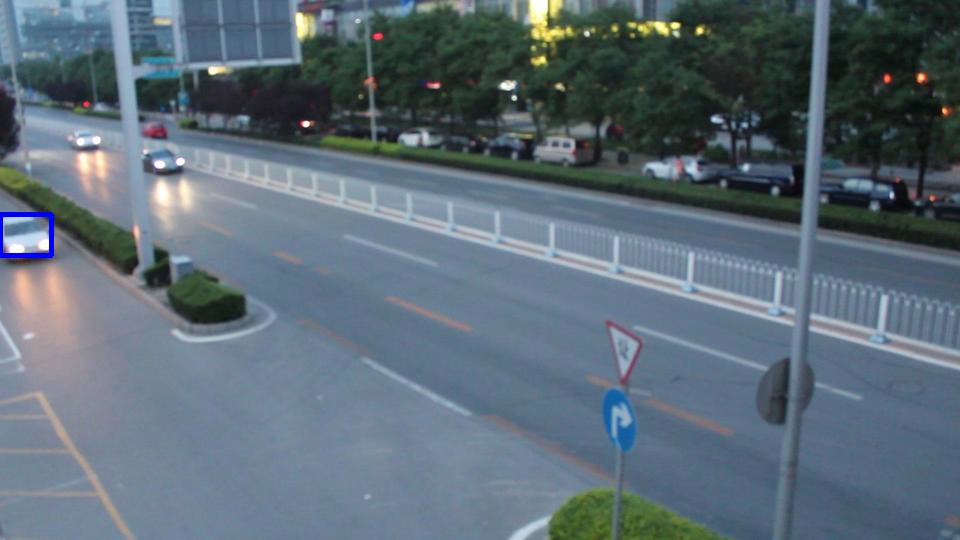}
      & \includegraphics[valign=c,width=0.2\linewidth]{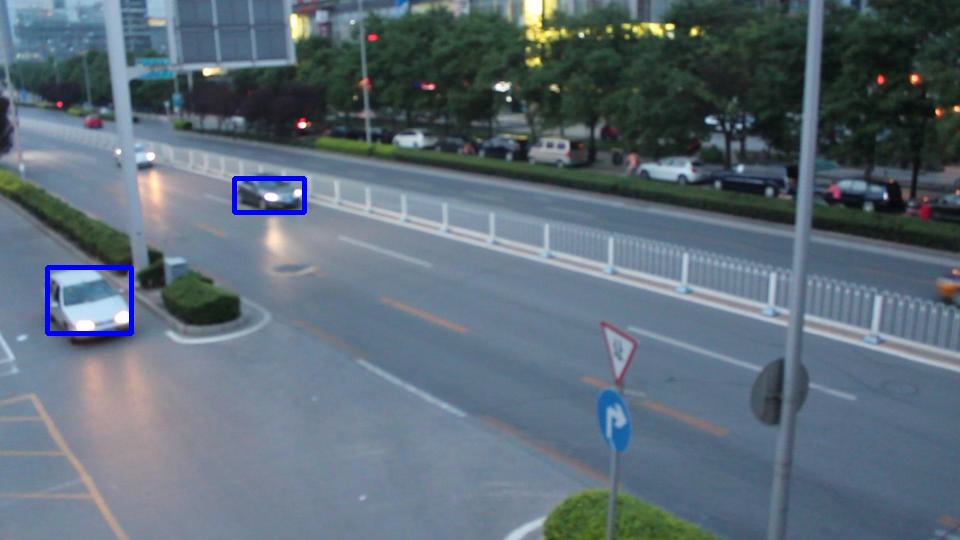}
      & \includegraphics[valign=c,width=0.2\linewidth]{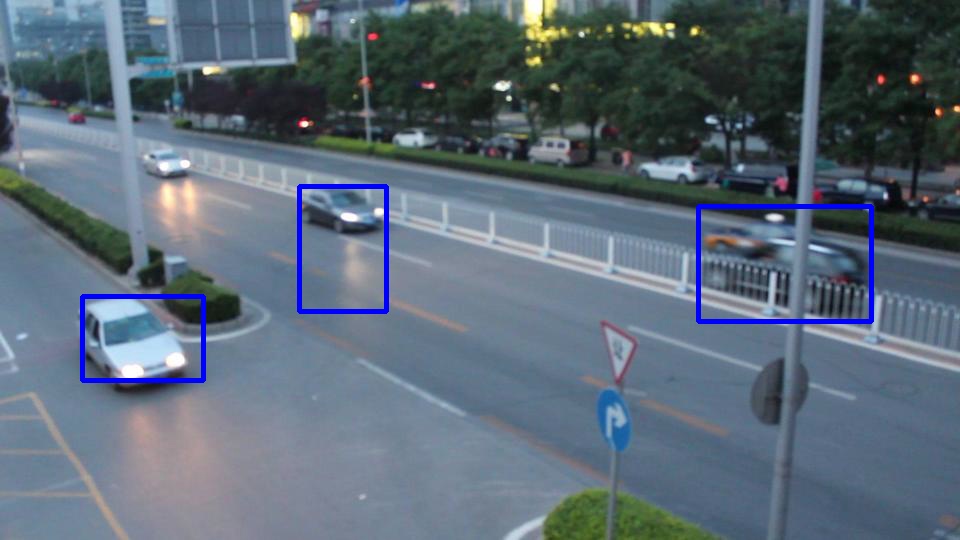}
      & \includegraphics[valign=c,width=0.2\linewidth]{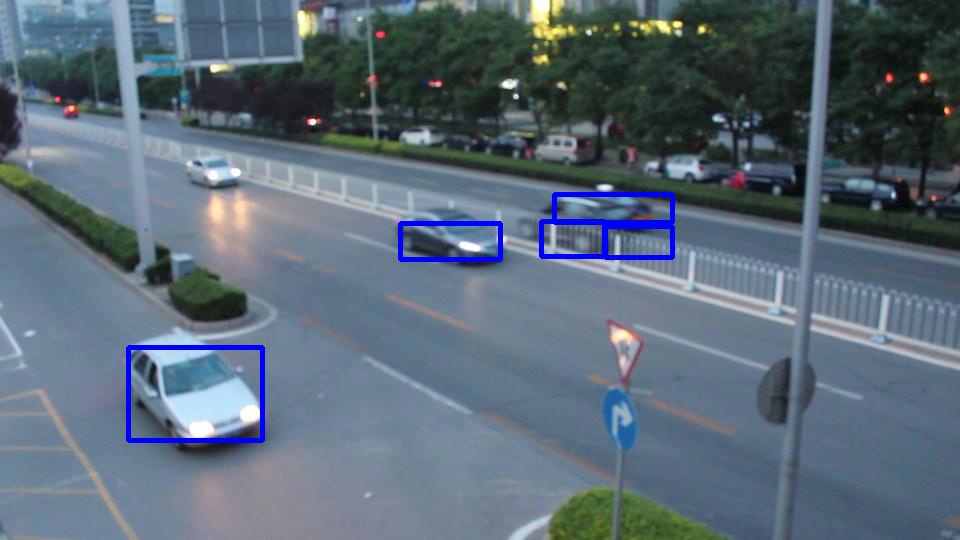}
  \end{tabular}
  \caption{Examples of extracted bounding boxes from supervised and unsupervised detections of road users in evaluated sequence.}
  \label{fig_seq}
\end{figure}

The UA-Detrac~\cite{wen2015ua} and UrbanTracker~\cite{jodoin2016tracking} datasets were used for the evaluation in this study because they include four challenging real-world traffic videos with 4 to 20 targets in the same frame simultaneously under different environmental conditions with varying types of annotated road users. The videos contain 600 to 1000 frames respectively. Evaluation of performances for the two datasets is performed using the standardized CLEAR MOT metrics \cite{bernardin2008evaluating}. Because unsupervised detections are less precise in their localization and extent (see Figure~\ref{fig_seq}), an intersection over union of 0.3 is used for computing the evaluation metrics as in previous work~\cite{beaupre2018improving}.
\subsection{Experimental setup for the UA-Detrac dataset}

Comparing supervised and unsupervised detections is not trivial because datasets are designed with one or the other in mind. UA-Detrac does not include annotations for pedestrians, bikes and motorcycles. Due to the nature of unsupervised methods in producing the input objects for the tracker, it is observed that the presence of these unannotated road users in the frames will severely affect the quality of inputs for tracking and perhaps good trajectories without corresponding annotations will be produced, but penalized in the MOTA. Hence, in order to allow for fair comparisons of performance for the two sources of inputs in the tracking phase, we have chosen 22 videos from the training set for this evaluation where there are no (or very few) pedestrians, bikes and motorcycles.
The videos are recorded at 25 fps (frame per seconds) with resolution of $960$x$540$ pixels.
The chosen videos include different angles of observations with varying illumination and weather conditions. Comparison of existing methods on the dataset is done by running the trackers on these videos individually to obtain their MOTA and MOTP results. 

For an unsupervised method, to get the detections, the background subtraction method typically observes the video for some time to learn the background. In UA-Detrac, objects have to be detected and tracked from the first frame of the video. Therefore, to simulate the normal condition in which an unsupervised method would be applied, for each selected video, $k$~frames are selected randomly over the whole video for learning the background. That way, foreground objects can then be detected from frame 1 in the tracking evaluation. Hence practically, for the unsupervised approach to work on this dataset, it has to ``see'' certain portion of the frames of the video before doing the actual foreground detection. Hence to allow fair comparison with the supervised methods, we are conducting experiment on videos of the training set, where the detector has ``seen'' the data as well. 

In practical applications where the evaluation is performed on new unseen data, it is expected that the tracking performance will be lower for both type of detectors because of some deterioration in quality of the detections obtained.

For the supervised detections as used in our method, RetinaNet with VGG-16 as backbone is trained on the training set of UA-Detrac. The detected objects with confidence lower than 0.4 are filtered out before tracking. As for unsupervised detections, only bounding boxes with areas that are greater than $2000$ pixels are allowed as input to the tracker. These steps are to ensure that only input objects that are valid in terms of size and confidence will be used for our MOT evaluation. Indeed, presence of spurious noise and incorrect detections can have a detrimental effect on the overall tracking performances. The supervised and unsupervised detections used in our experiments with UA-Detrac can be downloaded at this link (\url{https://github.com/HuiLee-Ooi/MF-Tracker}).

Besides comparing results of our proposed tracker with the different detection sources, tracking performances of existing trackers, under similar experimental settings with supervised and unsupervised detections, are presented as well in Table~\ref{tab_uadetrac}.

At the time of writing, the current reported three best trackers in the dataset official website are based on~\cite{bochinski2017high} with detection results from~\cite{wang2017evolving} and~\cite{girshick2014rich}. However, since we are not able to run the tracker version with the Kalman filter on the individual videos presented in this study (the public code does not work), only results of EB+IOU and RCNN+IOU are reported (as IoU + EB and IoU + RCNN in the table). 


\subsection{Experimental setup for the UrbanTracker dataset}
In this experiment, all four videos in the UrbanTracker dataset are used to evaluate and compare with existing methods. 

The optimal filter for the size of detections varies depending on video due to the inherently different scenarios. For a fair comparison, we are using the same parameter settings as presented by \cite{jodoin2016tracking}. 
Meanwhile, due to the limited amount of data in the dataset, supervised detection inputs are results of RetinaNet detection with VGG-16 backbone trained on the UA-Detrac training set. The confidence threshold for filtering out the input bounding boxes from supervised sources is set at $0.4$ for all videos. For unsupervised detections, extra frames are available before the annotated video segments. They are thus used to learn the background model.

The MOT performances for our proposed MF-Tracker (with supervised and unsupervised detections) compared to existing methods are presented in Table~\ref{tab_urbantracker}.

\begin{table*}[!htbp]
\centering
\caption{Comparison of MOTA and MOTP performances of trackers on the UrbanTracker dataset. For tracker names, the part following "+" indicates the method used to obtain the detections. \textbf{Boldface} indicates the best result,  \underline{Underline} indicates the second best result and \textit{{\color[HTML]{009901}Italicized green}} indicates the third best result. * indicates that the reported results are taken from original published works without re-running the methods. RL indicates Rene-Levesque and Sher. indicates Sherbrooke}
\label{tab_urbantracker} 
\begin{tabular}{c|cc|cc|cc|cc|cc}
\hline
\multirow{2}{*}{Video Seq.}                    & \multicolumn{6}{c|}{Unsupervised detections}                                                                                                                                                                                                                                   & \multicolumn{4}{c}{Supervised detections}                                                                                                                                                                         \\
                 & \multicolumn{2}{c|}{
                   MF-Tracker +} 
                   & \multicolumn{2}{c|}{
                   UrbanTracker +} 
                   & \multicolumn{2}{c|}{
                   MKCF +}  
                   & \multicolumn{2}{c|}{
                   MF-Tracker + } 
                   & \multicolumn{2}{c}{ Ooi et al \cite{ooi2018multiple}} \\
                
                   &\multicolumn{2}{c|}{
                   IMOT-PAWCS} 
                   & \multicolumn{2}{c|}{
                   IMOT * \cite{beaupre2018improving}} 
                   & \multicolumn{2}{c|}{
                   ViBe * \cite{yang2017multiple} }  
                   & \multicolumn{2}{c|}{
                   RetinaNet } 
                   & \multicolumn{2}{c}{+ RFCN \cite{dai2016r} } \\

 & MOTA                                   & MOTP                                                             & MOTA                                   & MOTP                                                          & MOTA                                  & MOTP  & MOTA                                                           & MOTP                                    & MOTA                                     & MOTP                                             \\
\hline
Rouen              & {\ul 0.5805}                           & {\color[HTML]{009901} \textit{0.6035}}                           & \textbf{0.670}                         & {\ul 0.620}                                                   & {\color[HTML]{009901} \textit{0.501}} & 0.582 & 0.133                                                         & 0.885                                  & -0.188                                  & \textbf{0.687}                                  \\
Sher.         & {\ul 0.609}                           & 0.5771                           & \textbf{0.690}                         & {\color[HTML]{009901} \textit{0.590 }}                                                        & 0.317                                 & 0.553 & {\color[HTML]{009901} \textit{0.3771}}                         & \textbf{0.915}                         & 0.027                                   & {\ul 0.7490}                                     \\
St-Marc            & {\ul 0.643}                           & {\color[HTML]{009901} \textit{0.6849}}                           & \textbf{0.653}                         & 0.682                                                         & {\color[HTML]{009901} \textit{0.463}} & 0.652 & 0.1124                                                         & \textbf{0.951}                         & -0.366                                  & {\ul 0.723}                                     \\
RL      & {\ul 0.3491}                           & {\ul 0.712}                                                     & \textbf{0.613}                         & {\color[HTML]{009901} \textit{0.705}}                         & {\color[HTML]{009901} \textit{0.334}} & 0.531 & 0.273                                                          & \textbf{0.901}                         & NA                                       & NA      \\                                        
\hline
\end{tabular}
\end{table*}

\section{Results}
For the UA-Detrac dataset, generally, the trackers with supervised detections give better tracking performances than the ones with unsupervised detections, as shown in Table~\ref{tab_uadetrac}. 
MF-Tracker outperformed all the compared methods when coupled with supervised detections.

Supervised detections on the UA-Detrac dataset work very well, where MF-Tracker + RetinaNet achieved a mean MOTA of 0.7638 and a mean MOTP of 0.8884, whereas unsupervised detections are not as good with MF-Tracker + IMOT-PAWCS only achieving mean MOTA of 0.2673 and mean MOTP of 0.6527, despite the use of a state-of-the-art background subtraction method. 

Despite the trend of supervised detectors overwhelmingly giving better performances than unsupervised detectors, it is interesting to note that for some videos (MVI\textunderscore40241, MVI\textunderscore40243 and MVI\textunderscore40244), our tracker with unsupervised detections ranked in third place, being quite competitive with the second ranked method (IoU + RCNN) that is based on supervised detections. These three videos are observed to have fast vehicles moving, causing motion blur. 
It is also observed that the use of state-of-the-art background subtraction (PAWCS~\cite{st2016universal}) with MKCF (Multiple Kernelized Correlation Filters) improves the performance of the original MKCF that uses ViBe background subtraction. Similarly, the use of PAWCS~\cite{st2016universal} with IMOT with our tracker has improved the tracking performance compared to the original implementation of the IMOT approach based on ViBe~\cite{barnich2010vibe}. IMOT post-process results from background subtraction by using optical flow and edges to solve object merging. 

On the contrary, the results on the UrbanTracker dataset are showing a different trend. Table~\ref{tab_urbantracker} shows that trackers with unsupervised detections give better performances in terms of multiple object tracking accuracy (MOTA), especially UrbanTracker + IMOT~\cite{beaupre2018improving}. Our proposed tracker with unsupervised detections (IMOT boxes from PAWCS background subtraction) ranked second in the comparison. However, it must be noted that the results reported in both~\cite{beaupre2018improving} and~\cite{yang2017multiple} are using parameters that are specifically tuned to each video in the dataset. In contrast, aside from the filter for input size in the tracker that varies according to video (which is a useful step given the disparity of target input size among the videos and because those filter sizes were used by the competing methods), the proposed MF-Tracker is applied with identical parameter settings for all the evaluated videos in the dataset. Still, MF-Tracker with unsupervised detections obtains competitive results with respect to \cite{beaupre2018improving, yang2017multiple} for Rouen, Sherbrooke and St-Marc, although tracking performance on Rene-Levesque is significantly worse.

\begin{table*}[!htbp]
\centering
\caption{Comparison of tracking results on videos from UrbanTracker dataset based on the different individual features}
\label{tab_ablation} 
\begin{tabular}{|c|cccccc|}
\hline
Features            & Correct Tracks & Misses & FP   & Mismatches & MOTP  & MOTA  \\ \hline
distance    & 19358          & 5491   & 5182 & 89         & 0.677 & 0.567 \\ 
color       & 19292          & 5557   & 5371 & 141        & 0.677 & 0.555 \\ 
label       & 18968          & 5881   & 5193 & 271        & 0.679 & 0.543 \\ 
REID       & 19090          & 5759   & 6761 & 654        & 0.678 & 0.470 \\ \hline
\end{tabular}
\end{table*}


Effects of the four different feature cost on our proposed tracker were studied individually on the UrbanTraccker dataset with supervised inputs in Table \ref{tab_ablation}, where number of correct tracks, misses, false positives (FP) and mismatches of the four videos are accumulated. It is observed that the compared features gives fairly similar MOTP (0.68) and MOTA ranges from 0.47 to 0.57. Spatial distance appears to be the best performing feature whereas REID is the worst performing feature. Therefore we used the following weights $\alpha =0.7, \beta=0.1, \gamma=0.1$ and $\lambda = 0.1$ in the experiments. 

\section{Discussion}
The quick impression from the presented results is that supervised methods give better detections for the UA-Detrac dataset and conversely, unsupervised detections work better on the UrbanTracker dataset. 

For UA-Detrac, while the use of state-of-the-art background subtraction might help improving the tracking results (comparing original MKCF with ViBe and MKCF with PAWCS), it is obvious that the methods with supervised detections are the clear winners. While one could argue that good results are expected since the videos are part of the training set, similar conditions can be said on the unsupervised methods as well since each video are ``seen'' to build the background model before producing foreground outputs (detections) for tracking purposes. However, despite a similar amount of learning on the data itself, methods with unsupervised detections with fixed parameter settings still yield poor results overall. 

Unsupervised object detection methods struggle with high density traffic were all objects become merged together. Supervised object detection methods handle these cases better because each road user is individually detected. Also, for unsupervised detection methods, in night conditions, car headlights generate foreground regions that are then tracked as ghost objects. They are ignored by supervised detection methods.

In any case, our proposed tracker with unsupervised detections (MF-Tracker + IMOT-PAWCS) is the best performing method among the methods with such detections, and it managed to rank third on three of the videos in terms of MOTA, effectively outperforming a method with supervised detections (IoU + EB). These videos are revealed to be containing high speed vehicles that appear slightly blurry in the frame, possibly causing the supervised detector to produce less accurate detections for the tracking framework. On the other hand, the camera that is statically positioned ensure that the backgrounds of the videos are properly learned without a lot of noise by the unsupervised detector, thereby producing detections of satisfactory quality to proceed with tracking. It must be noted that while the videos in the UA-Detrac dataset are taken from fixed camera setups, some inevitable environmental conditions such as windy weather can affect the quality of foreground given by unsupervised detectors as the camera is slightly moving and vibrating. In these cases, results show that newer methods (e.g. PAWCS) can better handle this issue than older methods (e.g.\ ViBe), where street markings and highways dividers are detected as objects. 

As we delve deeper in interpreting the results, it is observed that the supervised detectors do not perform as well on the UrbanTracker dataset as on the UA-Detrac dataset because the datasets contain inherently very different scenarios. The UA-Detrac dataset contains a large number of videos in similar locations and angles with subtle differences, such as illumination at different time of day. In contrast, the four videos in the UrbanTracker dataset are captured at entirely different locations and the different heights of installation of the cameras cause the captured objects in the frames to be highly varied in sizes and scales. UrbanTracker also contains a larger variety of viewpoints. The work of~\cite{ooi2018multiple} on UrbanTracker dataset has previously shown that a supervised detector performed poorly on the dataset, due to detector that produces too many false positive objects for tracking. Both MF-Tracker + RetinaNet and the tracker presented in~\cite{ooi2018multiple} are not trained on UrbanTracker itself due to the lack of available training videos. It is plausible that better results could be achieved by supervised detectors with more relevant training data, which is unfortunately lacking for proper training. 

The best performances on the UrbanTracker dataset are from UrbanTracker + IMOT~\cite{beaupre2018improving}, while our proposed tracker with unsupervised detections ranked second in terms of MOTA for all the videos. However, aside from the size filter for the unsupervised detections to be fed into the tracker, our proposed tracker retained all the same parameters and settings for all the videos. This is not the case for the tracker parameters in the works of~\cite{beaupre2018improving} and~\cite{yang2017multiple} that have been tuned to each of the specific videos in the dataset to achieve competitive final results. It is important to note that this could be the main reason why UrbanTracker + IMOT generally fare better on the UrbanTracker dataset. In practical real applications, however, it is desirable to have generic settings that is not overly tuned (overfit) to individual video sequence. 

\section{Conclusion}
We presented a novel multi-feature tracker (MF-Tracker) that comprises classical and modern features for the matching of objects across frames. In addition, we evaluated our tracker with either unsupervised or supervised object detection approaches to investigate their differences in MOT performance. Compared to the existing trackers evaluated on the datasets, our proposed tracker achieved the best performances on the UA-Detrac dataset and is highly competitive on the UrbanTracker dataset with fixed parameters for all videos during tracking. Supervised inputs, when sufficiently trained with available data, produce good inputs that lead to more accurate tracking of objects. Nevertheless, in simpler scenarios, if good training data is not available, unsupervised method can perform well and can be a good alternative that should not be neglected.

\subsubsection*{Acknowledgments}
This research is funded by FRQ-NT (Grant: 2016-PR- 189250) and Polytechnique Montr\'{e}al PhD Fellowship. The Titan X used for this research was donated by the NVIDIA Corporation. We acknowledge the contribution of Hughes Perreault for providing the RetinaNet detections.

\begin{landscape}
\begin{table}[!htbp]
\centering
\caption{Comparison of MOTA and MOTP performances of trackers with supervised and unsupervised detections on selected videos of UA-Detrac. For tracker names, the part following "+" indicates the method used to obtain detections. \textbf{Boldface} indicates best result,  \underline{Underline} indicates second best result and \textit{{\color[HTML]{009901}Italicized green}} indicates third best result.}
\label{tab_uadetrac}
 \begin{tabular}{c|cc|cc|cc|cc||cc|cc|cc}
 \multirow{3}{*}{Video Seq.}                   & \multicolumn{8}{c ||}{Unsupervised  detections}        & \multicolumn{6}{c}{Supervised detections}  
                   \\
                   \hline                                                                      
                   & \multicolumn{2}{c|}{} & \multicolumn{2}{c|}{} & \multicolumn{2}{c|}{MKCF +}& \multicolumn{2}{c||}{MF-Tracker+} & \multicolumn{2}{c|}{}           & \multicolumn{2}{c|}{}                                & \multicolumn{2}{c}{MF-Tracker +} 
                   \\
                                 & \multicolumn{2}{c|}{MKCF + ViBe}  & \multicolumn{2}{c|}{IMOT + ViBe}  & \multicolumn{2}{c|}{PAWCS}& \multicolumn{2}{c||}{IMOT-PAWCS} & \multicolumn{2}{c|}{IoU + EB}           & \multicolumn{2}{c|}{IoU + RCNN}                                & \multicolumn{2}{c}{ RetinaNet} \\
                   
\hline                   
                   
 & MOTA                 & MOTP                & MOTA                 & MOTP                & MOTA                    & MOTP                   & MOTA                                                                 & MOTP                                & MOTA                                   & MOTP            & MOTA                                   & MOTP                                   & MOTA                                                            & MOTP                                    \\
\hline
MVI\_39801 & -1.1280     & 0.5624     & -1.1493     & 0.5301     & 0.1309         & 0.5399        & 0.1970                                 & 0.5859 & {\color[HTML]{009901} \textit{0.6085}} & {\ul 0.8146}    & {\ul 0.6773}                           & {\color[HTML]{009901} \textit{0.7485}} & \textbf{0.8351}                        & \textbf{0.8536} \\
MVI\_39861 & -2.3416     & 0.5928     & -2.0680     & 0.5319     & -0.7905        & 0.5392        & -0.0244                                & 0.6201 & {\ul 0.7529}                           & {\ul 0.8423}    & {\color[HTML]{009901} \textit{0.5502}} & {\color[HTML]{009901} \textit{0.7312}} & \textbf{0.7824}                        & \textbf{0.8546} \\
MVI\_40191 & -1.1280     & 0.5624     & -0.6120     & 0.6227     & 0.1679         & 0.6001        & 0.3050                                 & 0.7227 & {\ul 0.7201}                           & {\ul 0.8979}    & {\color[HTML]{009901} \textit{0.5156}} & {\color[HTML]{009901} \textit{0.8337}} & \textbf{0.8549}                        & \textbf{0.9123} \\
MVI\_40192 & -1.6718     & 0.5357     & -1.0181     & 0.5948     & 0.3896         & 0.6055        & 0.3615                                 & 0.6915 & {\ul 0.5273}                           & {\ul 0.8807}    & {\color[HTML]{009901} \textit{0.4574}} & {\color[HTML]{009901} \textit{0.8145}} & \textbf{0.7999}                        & \textbf{0.8918} \\
MVI\_40201 & -2.4787     & 0.5489     & -0.7528     & 0.5861     & 0.4245         & 0.6182        & 0.3321                                 & 0.6687 & 0.4643                                 & \textbf{0.8897} & {\ul 0.6324}                           & {\color[HTML]{009901} \textit{0.8168}} & \textbf{0.8009}                        & {\ul 0.8873}    \\
MVI\_40204 & -0.9831     & 0.5456     & -0.7504     & 0.5700     & 0.2506         & 0.6050        & 0.2368                                 & 0.6651 & \textbf{0.7799}                        & {\ul 0.8645}    & {\ul 0.6676}                           & {\color[HTML]{009901} \textit{0.7647}} & 0.5312                                 & \textbf{0.8715} \\
MVI\_40211 & -5.1159     & 0.6144     & -3.0001     & 0.5892     & 0.1305         & 0.6642        & 0.3414                                 & 0.6514 & \textbf{0.8491}                        & {\ul 0.9011}    & {\color[HTML]{009901} \textit{0.6354}} & {\color[HTML]{009901} \textit{0.7703}} & {\ul 0.7019}                           & \textbf{0.9017} \\
MVI\_40212 & -3.3782     & 0.6059     & -1.9731     & 0.5858     & 0.0832         & 0.6576        & 0.2650                                 & 0.6438 & \textbf{0.8446}                        & \textbf{0.8952} & {\color[HTML]{009901} \textit{0.6485}} & {\color[HTML]{009901} \textit{0.7731}} & {\ul 0.7452}                           & {\ul 0.8841}    \\
MVI\_40213 & -3.1261     & 0.5969     & -1.7845     & 0.5928     & 0.2929         & 0.6594        & 0.3957                                 & 0.6430 & \textbf{0.8458}                        & \textbf{0.9023} & {\color[HTML]{009901} \textit{0.5389}} & {\color[HTML]{009901} \textit{0.7699}} & {\ul 0.7028}                           & {\ul 0.8920}    \\
MVI\_40241 & -0.5776     & 0.5802     & -0.3539     & 0.6214     & 0.3978         & 0.6246        & {\color[HTML]{009901} \textit{0.4493}} & 0.6880 & 0.3936                                 & {\ul 0.8998}    & {\ul 0.6279}                           & {\color[HTML]{009901} \textit{0.7821}} & \textbf{0.7535}                        & \textbf{0.9116} \\
MVI\_40243 & -0.0120     & 0.5995     & -0.0934     & 0.6424     & 0.3950         & 0.6177        & {\color[HTML]{009901} \textit{0.4828}} & 0.6862 & 0.2845                                 & {\ul 0.9009}    & {\ul 0.5216}                           & {\color[HTML]{009901} \textit{0.7860}} & \textbf{0.7695}                        & \textbf{0.9116} \\
MVI\_40244 & 0.1491      & 0.5771     & 0.0872      & 0.6458     & 0.4985         & 0.6091        & {\color[HTML]{009901} \textit{0.5448}} & 0.6770 & 0.1784                                 & {\ul 0.9044}    & {\ul 0.5818}                           & {\color[HTML]{009901} \textit{0.7843}} & \textbf{0.7316}                        & \textbf{0.9139} \\
MVI\_40752 & -0.5572     & 0.5888     & -0.3374     & 0.6301     & -0.0946        & 0.5952        & 0.2330                                 & 0.6586 & {\ul 0.6464}                           & \textbf{0.8799} & {\color[HTML]{009901} \textit{0.5782}} & {\color[HTML]{009901} \textit{0.7521}} & \textbf{0.7607}                        & {\ul 0.8788}    \\
MVI\_40871 & -0.4175     & 0.4671     & -0.5729     & 0.4176     & -0.1294        & 0.5151        & 0.0418                                 & 0.5522 & {\color[HTML]{009901} \textit{0.1642}} & {\ul 0.8861}    & {\ul 0.4538}                           & {\color[HTML]{009901} \textit{0.8061}} & \textbf{0.8208}                        & \textbf{0.9208} \\
MVI\_40962 & -0.4033     & 0.5940     & 0.0078      & 0.6796     & -0.2181        & 0.5943        & 0.3114                                 & 0.7298 & {\color[HTML]{009901} \textit{0.6969}} & {\ul 0.9140}    & {\ul 0.8478}                           & {\color[HTML]{009901} \textit{0.8488}} & \textbf{0.8696}                        & \textbf{0.9240} \\
MVI\_40963 & -0.3398     & 0.5442     & -0.0243     & 0.6428     & -0.1518        & 0.5237        & 0.2407                                 & 0.6730 & \textbf{0.7308}                        & \textbf{0.8637} & {\ul 0.7056}                           & {\color[HTML]{009901} \textit{0.7699}} & {\color[HTML]{009901} \textit{0.5945}} & {\ul 0.8441}    \\
MVI\_40981 & -0.6972     & 0.5346     & -0.4488     & 0.5951     & -2.0140        & 0.4758        & 0.0065                                 & 0.6000 & {\color[HTML]{009901} \textit{0.7529}} & {\ul 0.8915}    & {\ul 0.8122}                           & {\color[HTML]{009901} \textit{0.7932}} & \textbf{0.8964}                        & \textbf{0.9168} \\
MVI\_41063 & 0.1081      & 0.6198     & 0.1116      & 0.6271     & 0.4788         & 0.6399        & 0.3598                                 & 0.6507 & {\ul 0.7666}                           & {\ul 0.8738}    & {\color[HTML]{009901} \textit{0.7283}} & {\color[HTML]{009901} \textit{0.7868}} & \textbf{0.7939}                        & \textbf{0.8870} \\
MVI\_41073 & -1.1475     & 0.6197     & -0.6290     & 0.6440     & -0.4349        & 0.6355        & 0.2526                                 & 0.6820 & \textbf{0.8098}                        & \textbf{0.8889} & {\ul 0.7954}                           & {\color[HTML]{009901} \textit{0.7699}} & {\color[HTML]{009901} \textit{0.7320}} & {\ul 0.8732}    \\
MVI\_63552 & -3.2291     & 0.5804     & -2.2304     & 0.5329     & -0.1617        & 0.6421        & 0.1696                                 & 0.6249 & {\ul 0.6236}                           & {\ul 0.8334}    & {\color[HTML]{009901} \textit{0.5364}} & {\color[HTML]{009901} \textit{0.7487}} & \textbf{0.7518}                        & \textbf{0.8549} \\
MVI\_63553 & -3.0276     & 0.5699     & -1.5817     & 0.5530     & -0.1176        & 0.6256        & 0.1720                                 & 0.6161 & {\ul 0.7878}                           & {\ul 0.8455}    & {\color[HTML]{009901} \textit{0.5474}} & {\color[HTML]{009901} \textit{0.7433}} & \textbf{0.8032}                        & \textbf{0.8499} \\
MVI\_63554 & -3.0283     & 0.5657     & -1.9908     & 0.5631     & 0.0836         & 0.6122        & 0.2065                                 & 0.6278 & {\ul 0.7213}                           & \textbf{0.8739} & {\color[HTML]{009901} \textit{0.5668}} & {\color[HTML]{009901} \textit{0.7698}} & \textbf{0.7707}                        & {\ul 0.8660}    \\
\hline
average    & -1.5696     & 0.5730     & -0.9620     & 0.5908     & -0.0177        & 0.6000        & 0.2673                                 & 0.6527 & {\ul 0.6341}                           & {\ul 0.8793}    & {\color[HTML]{009901} \textit{0.6194}} & {\color[HTML]{009901} \textit{0.7802}} & \textbf{0.7638}                        & \textbf{0.8864}
\end{tabular}
\end{table}

\end{landscape}

\bibliographystyle{splncs04}
\bibliography{egbib}

\end{document}